\documentclass[letterpaper, 10 pt, conference]{ieeeconf}  

\IEEEoverridecommandlockouts                              

\usepackage{multicol}
\usepackage{multirow}
\usepackage{float} 
\usepackage{siunitx} 
\usepackage{tablefootnote}
\usepackage{adjustbox}
\usepackage{array}
\usepackage{booktabs}
\usepackage{makecell}

\usepackage{svg}
\usepackage{graphicx}
\usepackage{subcaption}

\usepackage{amsmath}
\usepackage{gensymb}
\usepackage{amssymb} 
\usepackage{relsize} 

\usepackage{cite} 

\makeatletter
\let\NAT@parse\undefined
\makeatother
\usepackage[hidelinks]{hyperref}

\usepackage[capitalize]{cleveref}
\crefname{figure}{Fig.}{Figs.}
\Crefname{figure}{Figure}{Figures}
\crefname{table}{Table}{Tables}
\Crefname{table}{Table}{Tables}

\title{\LARGE \bf
Are We Ready for Real-Time LiDAR Semantic Segmentation in Autonomous Driving?
}

\author{Samir Abou Haidar$^{1,2}$, Alexandre Chariot$^{1}$, Mehdi Darouich$^{1}$, Cyril Joly$^{2}$ and Jean-Emmanuel Deschaud$^{2}$
\thanks{$^{1}$The authors are with Paris-Saclay University, CEA, List, F-91120, Palaiseau, France
        {\tt\small first\_name.last\_name@cea.fr}}%
\thanks{$^{2}$The authors are with the Centre for Robotics, Mines Paris-PSL, 75006 Paris, France
        {\tt\small first\_name.last\_name@minesparis.psl.eu}}%
}

\begin{document}

\maketitle
\thispagestyle{empty}
\pagestyle{empty}

\begin{abstract}
Within a perception framework for autonomous mobile and robotic systems, semantic analysis of 3D point clouds typically generated by LiDARs is key to numerous applications, such as object detection and recognition, and scene reconstruction. Scene semantic segmentation can be achieved by directly integrating 3D spatial data with specialized deep neural networks. Although this type of data provides rich geometric information regarding the surrounding environment, it also presents numerous challenges: its unstructured and sparse nature, its unpredictable size, and its demanding computational requirements. These characteristics hinder the real-time semantic analysis, particularly on resource-constrained hardware architectures that constitute the main computational components of numerous robotic applications. Therefore, in this paper, we investigate various 3D semantic segmentation methodologies and analyze their performance and capabilities for resource-constrained inference on embedded NVIDIA Jetson platforms. We evaluate them for a fair comparison through a standardized training protocol and data augmentations, providing benchmark results on the Jetson AGX Orin and AGX Xavier series for two large-scale outdoor datasets: SemanticKITTI and nuScenes.


\end{abstract}

\section{INTRODUCTION}
\label{sec:introduction}
Autonomous mobile systems, including self-driving vehicles and mobile robots, are required to independently navigate in their environments. A fundamental prior step for autonomous maneuvering is to fully understand the surroundings and, more precisely, to analyze the semantics of the scenes. Such systems usually perceive the environment in the form of 3D point clouds that are generally obtained using LiDAR sensors. Hence, semantic segmentation has paramount importance in perception algorithms in terms of assigning each point of the 3D point cloud with a label corresponding to its class category (e.g., pedestrian, car, roads), and, therefore, providing fine-grained details of the surroundings. Presently, semantic segmentation is predominantly tackled using deep neural networks including~\cite{puy2023waffleiron, choy2019minkowski, tang2020spvnas,cortinhal2020salsanext,thomas2019kpconv, qi2017pointnet}, and serve as the foundational structures for other various 3D processes, such as 3D detection, thereby making it an integral component of perception algorithms and the center of our focus. Although very rich in information, 3D point sets present by nature  several challenges, such as an unstructured format, sparsity (presence of gaps and empty regions leading to incomplete scans), and size and complexity (contain a significant number of points). These characteristics make processing and understanding 3D point clouds computationally expensive, and rather challenging for real-time applications. 
\begin{figure}[H]
  \centering
    \includegraphics[width=0.475\textwidth]{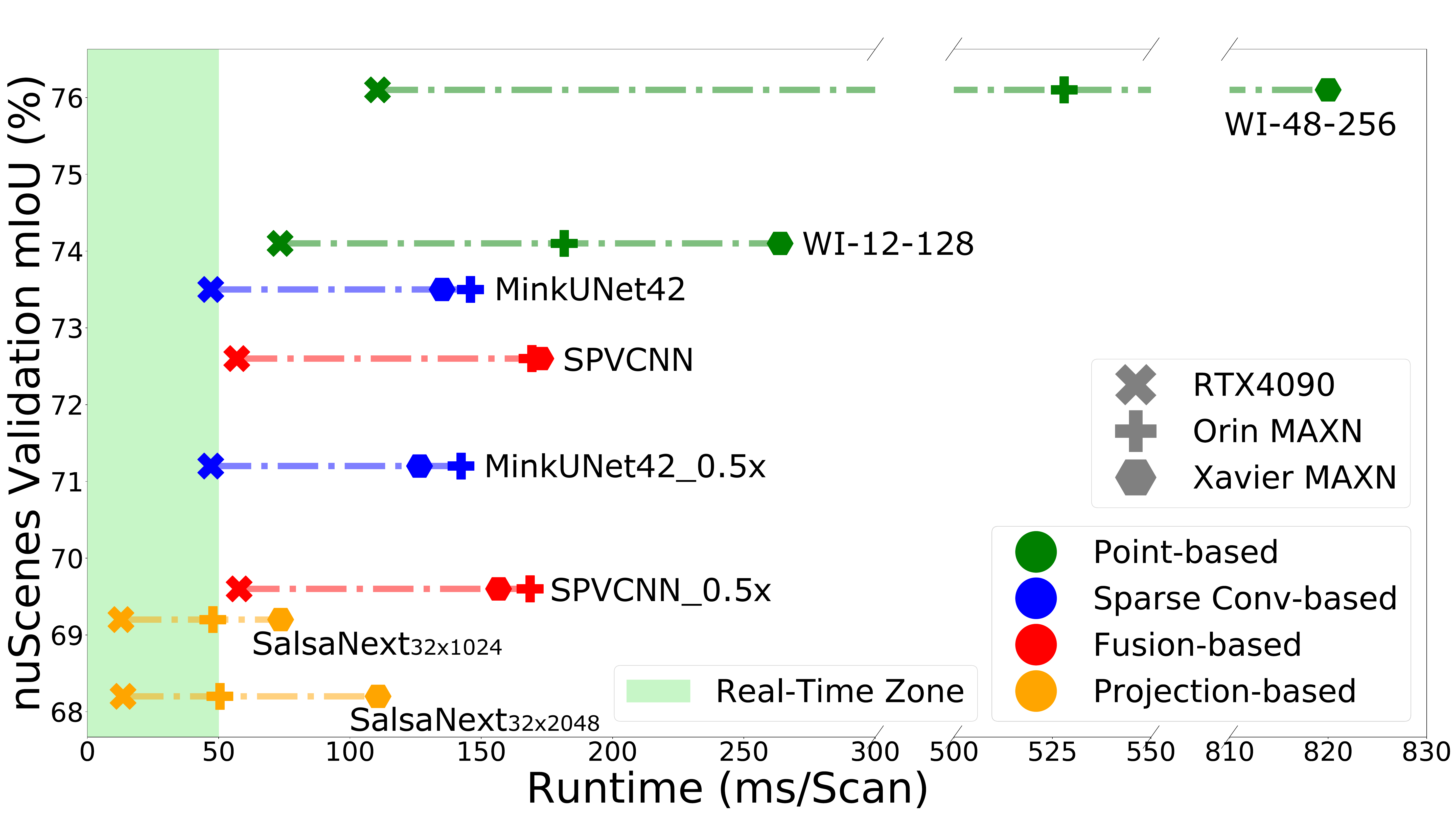}
    \caption{With a 20Hz acquisition sensor on nuScenes, only SalsaNext can be executed in real-time on RTX4090 and Jetson AGX Orin. Yet, its mIoU falls behind the other models which are far from real-time execution on Jetson platforms.}
    \label{fig:nuscenes}
\end{figure}


As safety is a top priority for autonomous mobile systems, it is required that 3D perception modules achieve high semantic segmentation accuracy and low latency at the same time, as such systems require real-time decision makings. However, hardware and computation resources in many robotic applications are constrained by the system form factor, power consumption, and heat dissipation, which makes it challenging for complex, and highly resource-dependent 3D semantic segmentation neural networks.

Further, hardware acceleration of embedded computing platforms has played a vital role in revolutionizing deep learning into practice. For example, NVIDIA Jetson platforms provide a recognized
performance for running neural networks for applications that involve image classification, and object detection, while being energy efficient, lightweight, and compact. Such
systems can be easily integrated in mobile autonomous and robotic systems provided that they can achieve good performance in real-time. A few previous studies have employed Jetson platforms for semantic segmentation on the AGX Xavier~\cite{lin2019edgesegnet}. Furthermore, existing benchmarks primarily address point cloud classification~\cite{ullah2020benchmarkingclassification}, object detection~\cite{choe2021benchmarkingobjectdetection}, or image semantic segmentation~\cite{siam2018comparativeimagesemseg}. Therefore, we provide in this work a benchmark of 3D point cloud semantic segmentation networks on the Jetson AGX Orin and AGX Xavier on large-scale outdoor datasets: SemanticKITTI~\cite{behley2019semantickitti} and nuScenes~\cite{caesar2020nuscenes}.

To ensure a fair benchmark, we develop a framework for neural networks on point clouds designed for high-end and embedded GPUs. In this framework, we reproduce all the networks and align their training protocols and data augmentations by adopting best practices from~\cite{qian2022pointnext, Xu2021rpvnet, puy2023waffleiron}. The framework is also tailored for seamless deployment on Jetson platforms, making it suitable for practical use cases. This paper primarily focuses on the benchmark results achieved through it, however, it will be released in the future.

\section{RELATED WORK}
\label{sec:relatedwork}

\subsection{3D Semantic Segmentation Methods}
\label{subsec:semanticsegmentationrelatedwork}

3D semantic scene labeling is vital for mobile autonomous and robotic systems. It involves classifying raw 3D points obtained from the system's sensors to provide fine-grained scene semantics. Related methods are usually divided into:


\textbf{Projection-based methods.} These methods are built on top of the image segmentation field (with a focus on CNN architectures used to segment RGB images) and target 3D point clouds acquired by rotating LiDARs. They project points onto different views for the purpose of working entirely on 2D feature maps~\cite{milioto2019rangenet++, zhang2020polarnet, cortinhal2020salsanext}.

\textbf{Point-based methods.}
Such approaches focus on directly processing points~\cite{qi2017pointnet, qian2022pointnext, puy2023waffleiron} instead of projecting onto intermediate representations. They are effective for dense input point clouds, but fall behind other methods for sparse outdoor ones, apart from~\cite{puy2023waffleiron} that reveals impressive capabilities in predicting sparse outdoor points. Recently, there have been a shift towards transformer-based architectures~\cite{wu2022ptv2, wu2024ptv3}.


\textbf{Sparse convolution-based methods.} 
 Various works exploit the sparsity in point clouds to perform sparse convolutions that discard entirely empty point cloud regions in the computation, thereby reducing memory consumption and computational expenses~\cite{choy2019minkowski, cheng2021(af)2-s3net, Hou2022pvkd} while preserving the effectiveness of output predictions.



\textbf{Fusion-based methods.} To improve semantic segmentation, such methods either combine the advantages of different point representations and views from a single input (3D LiDAR data)~\cite{tang2020spvnas, Xu2021rpvnet} or fuse different inputs from different sensors (camera and LiDAR)~\cite{Yan20222dpass}, and benefit from their different advantages. 




\subsection{Embedded 3D Applications on Jetson Platforms}
\label{subsec:embeddedrelatedwork}

Some works target 3D point cloud applications on embedded Jetson devices. For instance, \cite{ullah2020benchmarkingclassification} benchmarks Jetson platforms, including Nano, TX1, and AGX Xavier, for 3D point cloud classification, but only examines PointNet~\cite{qi2017pointnet} architecture on ModelNet-40, a 3D object-based synthetic and small dataset. Similarly, \cite{choe2021benchmarkingobjectdetection} analyzes deep learning-based 3D object detectors on Jetson platforms (AGX Xavier and Nano) by evaluating different YOLO versions.
Additionally, \cite{heitzinger2023fastunified} presented a lightweight system for real-time 3D object detection and tracking and evaluated it on the Jetson Orin Nano. Thus, the implementation of various state-of-the-art 3D semantic segmentation methods on embedded computing platforms has remained a challenge, and thus the primary focus of our work. We target two Jetson platforms, AGX Orin and AGX Xavier presented in \cref{tab:nvidia_jetson}, as they provide a balanced compromise between performance, efficiency, and cost, with varying levels of performance, memory, and power, making them suitable for benchmarking.


\begin{table}[t!]
\caption{The targeted Jetson platforms specifications.}
\label{tab:nvidia_jetson}
\centering
\resizebox{0.47\textwidth}{!}{
\begin{tabular}{|c|c|c|}
\hline
\textbf{Version} & \textbf{Jetson AGX Orin} & \textbf{Jetson AGX Xavier} \\
\hline
\textbf{Performance} & 275 TOPS & 32 TOPS \\
\hline

\textbf{GPU} & \begin{tabular}{@{}c@{}}2048-core NVIDIA \\ Ampere architecture GPU \\ with 64 Tensor Cores\end{tabular} & \begin{tabular}{@{}c@{}}512-core NVIDIA \\ Volta architecture GPU \\ with 64 Tensor Cores\end{tabular} \\
\hline

\textbf{CPU} & \begin{tabular}{@{}c@{}}12-core Arm® \\ Cortex®-A78AE v8.2 64-bit \\ CPU 3MB L2 + 6MB L3\end{tabular} & \begin{tabular}{@{}c@{}}8-core NVIDIA Carmel \\ Arm®v8.2 64-bit CPU \\ 8MB L2 + 4MB L3\end{tabular} \\
\hline

\textbf{Memory} & 64 GB & 32 - 64 GB \\
\hline

\textbf{Power} & 15W - 60W & 10W - 30W \\
\hline

\end{tabular}
}

\end{table}
\section{BENCHMARK SETUP}
\label{sec:benchmark}

\subsection{Datasets}
\label{subsec:datasets}
We focus on the semantic segmentation task for real-time applications\textemdash that is mobile autonomous and robotic systems; thus, we target two large-scale datasets to study outdoor semantic segmentation on SemanticKITTI~\cite{behley2019semantickitti} and nuScenes~\cite{caesar2020nuscenes} for the challenging perception application of autonomous vehicles in real-time. The point cloud acquisition sensors for those outdoor datasets are two different LiDAR scanners that provide different point cloud characteristics in terms of density/sparsity and total number of points.

\subsection{Selected 3D Neural Networks}
\label{subsec:neuralnetworks}
From each category presented in \cref{subsec:semanticsegmentationrelatedwork} we select a method and create a lighter variant from its baseline model for performance gains on embedded systems, and assess both models' performance on resource-constrained platforms:

From the projection-based methods, SalsaNext~\cite{cortinhal2020salsanext} provides a good trade-off between segmentation accuracy and runtime when compared to other methods from this category; it is developed based on the LiDAR range view image obtained from the spherical projection~\cite{milioto2019rangenet++}, which is the baseline of all projection-based methods.

We also consider WaffleIron~\cite{puy2023waffleiron} from the point-based methods, a recent backbone that solely relies on dense 2D convolutions and generic MLPs, making it less complicated in terms of design choices than numerous other methods. It also proved to be among the best performing in this category on SemanticKITTI~\cite{behley2019semantickitti} and  nuScenes~\cite{caesar2020nuscenes}. Therefore, it is worth investigating on embedded systems.

From the sparse convolution-based methods, Minkowski~\cite{choy2019minkowski} proposed the most generalized high-dimensional sparse tensor computations that leverage the sparsity in point clouds to lower memory consumption and accelerate inference, while considering 3D convolutions instead of 2D ones that suffer from loss in geometry and topology. Other methods build on top of Minkowski~\cite{cheng2021(af)2-s3net, tang2020spvnas}, thereby making it suitable for benchmarking. For example, SPVConv~\cite{tang2020spvnas} incorporated the sparse 3D convolution with a point-wise MLP to subsequently fuse the extracted features from both levels together, to better classify small instances and increase the overall accuracy. This fusion-based method is also investigated.

\subsection{Experimental Setup}
\label{subsec:experimentalsetup}
We first train the models on NVIDIA GeForce RTX4090, and then only test their inference on the embedded systems.

For WaffleIron~\cite{puy2023waffleiron}, the model is denoted as WI-$L$-$F$, with $L$ being the total number of layers and $F$ the dimension of point tokens generated from the embedding layer. This embedding layer takes as input the low-level readily available features at each point\textemdash e.g, LiDAR intensity, height and range\textemdash and uses $k=16$ neighbors to merge global and local information around them to finally provide a point cloud with an $F$-dimensional token associated with each point. The obtained tokens will be updated $L$ times by a series of token-mixing and channel-mixing (a shared MLP across each point) layers, with the core components of the token-mixing part being a 2D projection along a main axis, a features discretization on a 2D grid, and a feed-forward network established with 2D dense convolutions. We adopt the following configurations: $(x,y)$ projection at all layers on a 2D grid of cell size $\rho = 40\ cm$ for SemanticKITTI , and a baseline sequence of $(x,y)$, $(x,z)$, and $(y,z)$ projection at layer $l=1$, $l=2$, and $l=3$, respectively, repeated until reaching $l=L$ on a 2D grid of cell size $\rho = 60\ cm$ for nuScenes. We assess the performance of two models of WaffleIron, namely \textbf{WI-48-256} and \textbf{WI-12-128}.

With regard to Minkowski~\cite{choy2019minkowski}, we use MinkowskiUNet42 (\textbf{MinkNet}), a widely recognized architecture and prominently referred to in~\cite{tang2020spvnas,cheng2021(af)2-s3net,Hou2022pvkd,puy2023waffleiron} for 3D semantic segmentation. We specify the dimensions of the stem, encoder, and decoder channels to 32, [32, 64, 128, 256], and [256, 128, 96, 96], respectively. We set the voxel size to 5 cm on all datasets. Similarly, \textbf{SPVCNN}~\cite{tang2020spvnas} is built upon MinkowskiUNet42, by wrapping its residual sparse convolution blocks with a high-resolution point-based branch. We train two models of each network, with the second model obtained by pruning 50\% of all input and output channels of the 3D sparse convolutions at every layer, that is respectively \textbf{MinkNet\_0.5$\times$} and \textbf{SPVCNN\_0.5$\times$}. 


As for SalsaNext~\cite{cortinhal2020salsanext} (\textbf{SN}), it first applies a spherical projection of the point cloud by converting each point $p_i = (x,y,z)$ via a mapping $\mathbb{R}^3 \rightarrow \mathbb{R}^2$ to spherical coordinates, and then to image coordinates, thereby generating a LiDAR's native range view image, 
with specified desired height and width $(h,w)$ of the image representation. Following~\cite{cortinhal2020salsanext} when performing the projection, we store the 3D point coordinates $(x, y, z)$, the intensity value $(i)$, and the range $(r)$ as distinct image channels. This results in an image with dimensions $[w \times h \times 5]$, which is then fed to the same network originally designed in~\cite{cortinhal2020salsanext}. We train this network with two variants of the image size $[2048 \times 64 \times 5]$ and $[1024 \times 64 \times 5]$ on SemanticKITTI, and $[2048 \times 32 \times 5]$ and $[1024 \times 32 \times 5]$ on nuScenes, according to their corresponding sensor's FoV.

\subsection{Training Protocol}
\label{subsec:trainingprotocol}
All models are reproduced and retrained through our framework on a single NVIDIA GeForce RTX4090 GPU with 24 GB of memory using AdamW optimizer for 45 epochs on SemanticKITTI~\cite{behley2019semantickitti} and nuScenes~\cite{caesar2020nuscenes}, with a weight decay of 0.003, and a batch size of 3 (due to GPU memory capacity). For the learning rate, we use a scheduler with a linear warmup phase from 0 to 0.001 along the first 4 epochs and then a cosine annealing phase is used to decrease it to $10^{-5}$ at the end of the last epoch. The objective loss function used is a sum of the cross-entropy and the Lov\'{a}sz loss. To prevent overfitting, we implement classical point cloud augmentations: random rotation around the z-axis, random flip of the direction of the x-axis and y-axis, and random re-scaling. Following~\cite{puy2023waffleiron, Xu2021rpvnet, Yan20222dpass}, we implement instance cutmix on SemanticKITTI~\cite{behley2019semantickitti}, specifically targeting rare-class objects in order to enhance segmentation performance on this dataset. The approach involves extracting instances from the following classes: bicycle, motorcycle, person, bicyclist, and other vehicles. During training, we randomly select up to 40 instances from each of the selected classes and apply various random transformations, including rotation around the z-axis, flipping along the x or y axes, and random re-scaling for each instance. These instances are then placed at random locations on roads, in parking areas, or on sidewalks.

\subsection{Evaluation Metrics}
To qualitatively evaluate the accuracy of the methods in semantic predictions on different datasets, we use the mean Intersection-over-Union (mIoU) over all evaluation classes.

In addition, we assess the performance of these methods on the Jetson platforms by profiling them on the same group of scans from each dataset and report their pre-processing and post-processing times, their model inference time, and their RAM and GPU memory consumption on different power modes. We report the results on the MAXN power mode (representing the limits of each Jetson platform without any power budget) in \cref{sec:results}.  We omit the recordings on the very first scan to allow for GPU warm-up and prevent initial overheads on the CPU. We also evaluate each model's complexity in terms of the total number of parameters and the mean of multiply-accumulate operations (MACs) and study their effects on the overall performance. We also examine the actual power consumption on the Jetson platforms by using the \textit{tegrastats} utility, a powerful tool that provides statistics for Tegra-based devices. We measure the GPU and CPU power usage and report the total consumption as their sum.




\begin{table*}[t]
    \captionsetup{skip=0.1\baselineskip} 
    \vspace{0.2cm}
    \caption{Models' complexity evaluated over 1000 scans from the validation set of SemanticKITTI and nuScenes using a batch size of 1. *Values are directly obtained from~\cite{tang2020spvnas}.} 
    \label{tab:complexity}
    \centering
    \resizebox{0.88\textwidth}{!}{
    
    \begin{tabular}{|>{\scriptsize}c|>{\scriptsize}c|>{\scriptsize}c|>{\scriptsize}c|>{\scriptsize}c|>{\scriptsize}c|>{\scriptsize}c|>{\scriptsize}c|}
    \hline
    \multicolumn{4}{|>{\scriptsize}c|}{\textbf{Dataset}} & \multicolumn{2}{>{\scriptsize}c|}{\textbf{SemanticKITTI~\cite{behley2019semantickitti}}} & \multicolumn{2}{>{\scriptsize}c|}{\textbf{nuScenes~\cite{caesar2020nuscenes}}}  \\
    
    \multicolumn{4}{|>{\scriptsize}c|}{\textbf{\#Points/Scan} (Mean $\pm$ Std Dev)} & \multicolumn{2}{>{\scriptsize}c|}{(90,465 $\pm$ 6,306)} & \multicolumn{2}{>{\scriptsize}c|}{(22,180 $\pm$ 2,090)} \\
    \hline
    \hline
    
    \textbf{Model} & \textbf{Year} & \textbf{Category} & \textbf{\#Params (M)} & \textbf{\textbf{\#MACs (G)}} & \textbf{mIoU} & \textbf{\textbf{\#MACs (G)}} & \textbf{mIoU} \\
    
    \hline
    \hline
    
    \textbf{MinkNet~\cite{choy2019minkowski}} & \multirow{2}{*}{2019} & \multirow{2}{*}{Sparse Conv-based} & 21.7 & 113.9* & 64.3 & 22.2* & 73.5\\
    \cline{1-1}\cline{4-8}
    
    \textbf{MinkNet\_0.5$\times$~\cite{choy2019minkowski}} &  &  & 5.4 & 28.5 & 63.2 & 5.5 & 71.2 \\
    \cline{1-8}

    \textbf{SN\_$\rm{h}\times$2048~\cite{cortinhal2020salsanext}} & \multirow{2}{*}{2020} & \multirow{2}{*}{Projection-based} & 6.7 & 62.8 & 55.9 & 31.4 & 68.2 \\
    \cline{1-1}\cline{4-8}
    
    \textbf{SN\_$\rm{h}\times$1024~\cite{cortinhal2020salsanext}} &  &  & 6.7 & 31.4 & 54.4 & 15.7 & 69.2 \\
    \cline{1-8}
    
    \textbf{SPVCNN~\cite{tang2020spvnas}} & \multirow{2}{*}{2020} & \multirow{2}{*}{Fusion-based} & 21.8 & 118.6* & 65.3 & 23.1* & 72.6 \\
    \cline{1-1}\cline{4-8}
    
    \textbf{SPVCNN\_0.5$\times$~\cite{tang2020spvnas}} &  &  & 5.5 & 30 & 63.4 & 5.8 & 69.6 \\
    \cline{1-8}

    \textbf{WI-48-256~\cite{puy2023waffleiron}} & \multirow{2}{*}{2023} & \multirow{2}{*}{Point-based} & 6.8 & 457.1 & 65.8 & 122.4 & 76.1 \\
    \cline{1-1}\cline{4-8}
    
    \textbf{WI-12-128~\cite{puy2023waffleiron}} &  &  & 0.5 & 43.6 & 63.6 & 11.6 & 74.1  \\
    \cline{1-8}

    
    \hline

    \end{tabular}
    }

\end{table*}
\section{BENCHMARK RESULTS AND DISCUSSIONS}
\label{sec:results}


\subsection{General Overview}

~\cref{tab:complexity} reveals that largely reducing the complexity in terms of \#Params and \#MACs yields lighter models without substantial sacrifices in prediction accuracy (mIoU). Yet, this only reduces the Inference time and has no influence on the Pre-Processing phase. However, the amount of reduction is not always correlated, as a 90\% decrease in both \#Params and \#MACs from WI-48-256 to WI-12-128 results in a decrease of 85\%--86\%  in Inference time on SemanticKITTI, and  70\%--83\% on nuScenes across the different computing architectures as depicted in~\cref{tab:performance_metrics}. Yet, a 75\% decrease in \#Params and \#MACs from MinkNet to MinkNet\_0.5$\times$ only results in a decrease of 2\% in Inference time on the RTX4090, 22.9\% on AGX Orin, and 37.9\% on AGX Xavier on SemanticKITTI and less than 8\% on nuScenes for all hardware architectures. A similar trend is also observed when reducing SPVCNN to SPVCNN\_0.5$\times$. As for SalsaNext, a 50\% \#MACs reduction from SN\_64$\times$2048 to SN\_64$\times$1024 results in almost a 50\% decrease in Inference time, particularly for SemanticKITTI, but it is less significant for nuScenes.



Examining \cref{fig:nuscenes}, \cref{fig:semantickitti} and \cref{tab:performance_metrics} reveals that none of the evaluated models yield satisfactory outcomes in terms of total runtime on the embedded Jetson platforms apart from SalsaNext that can run in or near real-time on the AGX Orin and is the closest to achieving real-time execution on the AGX Xavier; hence, it is the most efficient in terms of semantic segmentation on the considered computing architectures in comparison to WaffleIron, Minkowski, and SPVCNN. However, SalsaNext falls behind all other methods in terms of segmentation accuracy on both datasets.

In terms of power consumption, an analysis from \cref{tab:performance_metrics} indicates that the GPU consistently draws more power than the CPU across all models on Jetson platforms. Notably, sparse convolutions in Minkowski and SPVCNN demonstrate increased efficiency, particularly on AGX Orin, thereby showcasing its high performance for sparse models due to its sparsity optimized Tensor cores. Additionally, reducing model complexity in \#Params and \#MACs leads to an overall decrease in power consumption, which is primarily attributed to mitigated GPU power draw. With no power budget (MAXN), both Jetson platforms never reach their maximal theoretical power, as indicated in \cref{tab:nvidia_jetson}.

\subsection{Pre-Processing Time}
The Pre-Processing phase includes all prerequisite CPU processes to prepare each input scan for network inference on the GPU. It also encompasses Post-Processing; but, since the latter is less significant, we adopt the Pre-Processing naming convention. Numerous works including \cite{puy2023waffleiron, behley2019semantickitti, zhang2018efficientconv}, assess the runtime of a neural network by only considering its Inference time. For example, \cite{zhang2018efficientconv} ignores the time needed in the initial voxelization step and the final step of projecting voxel predictions back to the initial point cloud. Similarly, \cite{puy2023waffleiron} compares several backbones, including WaffleIron, Minkowski, and SPVCNN in terms of Inference time only and demonstrates that WaffleIron is comparable to the other backbones on nuScenes and is slightly slower ($\times 1.7$) than Minkowski on SemanticKITTI. Contrary to this, we showcase, in addition to the Inference time, the importance of the Pre-Processing time. \cref{tab:performance_metrics} depicts that the Pre-Processing time of WaffleIron has a greater significance than all other models on every dataset. For instance, it is evaluated to be $\times 10.5$, $\times 8.4$, $\times 7.6$ more than that of Minkowski in the RTX4090, AGX Orin, and AGX Xavier on SemanticKITTI respectively; a similar trend is observed on nuScenes. Furthermore, it is observed that for WI-12-128, the Pre-Processing time surpasses the Inference time across every considered computing architecture and dataset. Additionally, SalsaNext's Pre-Processing duration constitutes a substantial portion, ranging from 35\% to 83\% of its overall runtime. Consequently, the significance of Pre-Processing time is notable, as it largely contributes to the Total Runtime and is crucial when assessing a model's real-time execution.

Given that the Pre-Processing phase is mainly carried out on the CPU, whereas the Inference occurs on the GPU, there exists an opportunity to mitigate the overall system latency through a pipeline approach, by overlapping input data Pre-Processing and Inference computations, which can enhance the system efficiency and potentially reduce the time required for completing the entire computational workflow.

\begin{table*}[t] 
\captionsetup{skip=0.1\baselineskip} 
\vspace{0.2cm}
  \caption{Performance metrics evaluated over 1000 scans from the SemanticKITTI and nuScenes validation sets using a batch size of 1. Mean Total Runtime values are presented with a standard deviation not exceeding 10\%. Peak GPU memory is reported, with RAM usage not exceeding 45 MB on SemanticKITTI and 25 MB on nuScenes for all models.}

  \label{tab:performance_metrics}
  \centering
  \resizebox{\textwidth}{!}{
  
  \begin{tabular}{|c|c|c|r@{ }c@{}r@{ }c@{}r@{}l|r@{ }r@{}c@{ }c@{}r@{}l|r@{ }c@{}r@{ }c@{}r@{}l|c|c|c|r@{ }r@{}c@{ }c@{}r@{}l|r@{ }c@{}r@{ }c@{}r@{}l|}
    \cline{3-36}
    \multicolumn{2}{c|}{} & \textbf{\textbf{mIoU}} & \multicolumn{18}{c|}{\textbf{Total Runtime} (Pre-Processing + Inference) \textbf{(ms)}} & \multicolumn{3}{c|}{\textbf{GPU Memory (MB)}} & \multicolumn{12}{c|}{\textbf{Total Power} (CPU + GPU) \textbf{(W)}}\\
    
    \cline{4-36}
    \multicolumn{2}{c|}{} & \textbf{\%} & \multicolumn{6}{c|}{\textbf{RTX4090}} & \multicolumn{6}{c|}{\textbf{Orin MAXN}}  & \multicolumn{6}{c|}{\textbf{Xavier MAXN}} & \textbf{RTX} & \textbf{Orin}  & \textbf{Xavier} & \multicolumn{6}{c|}{\textbf{Orin MAXN}}  & \multicolumn{6}{c|}{\textbf{Xavier MAXN}}\\
    
    \hline
    \multirow{8}{*}{\rotatebox[origin=c]{90}{\begin{tabular}{@{}c@{}}\textbf{SemanticKITTI} \\ \textbf{Validation~\cite{behley2019semantickitti}}\end{tabular}}} & \textbf{MinkNet~\cite{choy2019minkowski}} & 64.3 & \textbf{71} & ( & 22 & + & 49 & ) & \textbf{211} & ( & 45 & + & 166 & ) & \textbf{332} & ( & 66 & + & 266 & ) & 467 & 542 & 542 & \textbf{20.2} & ( & 2.6 & + & 17.6 & ) & \textbf{18.8} & ( & 2.5 & + & 16.3 & )\\
    \cline{2-36}
    & \textbf{MinkNet\_0.5$\times$~\cite{choy2019minkowski}} & 63.2 & \textbf{70} & ( & 22 & + & 48 & ) & \textbf{171} & ( & 43 & + & 128 & ) & \textbf{223} & ( & 58 & + & 165 & ) & 194 & 266 & 266 & \textbf{16.6} & ( & 2.8 & + & 13.8 & ) & \textbf{15.8} & ( & 2.9 & + & 12.9 & )\\
    
    \cline{2-36}
    & \textbf{SN\_64$\times$2048~\cite{cortinhal2020salsanext}} & 55.9 & \textbf{36} & ( & 28 & + & 8 & ) & \textbf{109} & ( & 58 & + & 51 & ) & \textbf{218} & ( & 87 & + & 131 & ) & 498 & 471 & 365 & \textbf{20.6} & ( & 3.4 & + & 17.2 & ) & \textbf{19.7} & ( & 3.0 & + & 16.7 & )\\
    \cline{2-36}
    & \textbf{SN\_64$\times$1024~\cite{cortinhal2020salsanext}} & 54.4 & \textbf{30} & ( & 25 & + & 5 & ) & \textbf{84} & ( & 58 & + & 26 & ) & \textbf{147} & ( & 80 & + & 67 & ) & 331 & 320 & 198 & \textbf{17.1} & ( & 3.8 & + & 13.3 & ) & \textbf{17.3} & ( & 3.8 & + & 13.5 & )\\

    \cline{2-36}
    & \textbf{SPVCNN~\cite{tang2020spvnas}} & 65.3 & \textbf{85} & ( & 26 & + & 59 & ) & \textbf{252} & ( & 43 & + & 209 & ) & \textbf{389}   & ( & 60 & + & 329 & ) & 585 & 657 & 657 & \textbf{20.2} & ( & 2.5 & + & 17.7 & ) & \textbf{18.2} & ( & 2.2 & + & 16.0 & )\\
    \cline{2-36}
    & \textbf{SPVCNN\_0.5$\times$~\cite{tang2020spvnas}} & 63.4 & \textbf{77} & ( & 21 & + & 56 & ) & \textbf{203} & ( & 43 & + & 160 & ) & \textbf{276} & ( & 61 & + & 215 & ) & 270 & 343 & 343 & \textbf{17.1} & ( & 2.8 & + & 14.3 & ) & \textbf{16.1} & ( & 2.7 & + & 13.4 & )\\

    \cline{2-36}
     & \textbf{WI-48-256~\cite{puy2023waffleiron}} & 65.8 & \textbf{370} & ( & 232 & + & 138 & ) & \textbf{1,847} & ( & 381 & + & 1,466 & ) & \textbf{3,010} & ( & 505 & + & 2,505 & ) & 1,276 & 2,384 & 2,378 & \textbf{20.7} & ( & 1.5 & + & 19.2 & ) & \textbf{15.1} & ( & 1.3 & + & 13.8 & ) \\
    \cline{2-36}
    & \textbf{WI-12-128~\cite{puy2023waffleiron}} & 63.6 & \textbf{251} & ( & 231 & + & 20  & ) & \textbf{565} & ( & 358 & + & 207 & ) & \textbf{840} & ( & 508 & + & 332 & ) & 629 & 1,186 & 1,178 & \textbf{14.1} & ( & 2.3 & + & 11.8 & ) & \textbf{11.9} & ( & 2.3 & + & 9.6 & ) \\
    
    \hline
    \hline
    
    \multirow{8}{*}{\rotatebox[origin=c]{90}{\begin{tabular}{@{}c@{}}\textbf{nuScenes} \\ \textbf{Validation~\cite{caesar2020nuscenes}}\end{tabular}}}   & \textbf{MinkNet~\cite{choy2019minkowski}} & 73.5 & \textbf{47} & ( & 8 & + & 39 & ) & \textbf{146} & ( & 22 & + & 124 & ) & \textbf{135} & ( & 34 & + & 101 & ) & 322 & 351 & 351 & \textbf{10.6} & ( & 2.8 & + & 7.8 & ) & \textbf{13.8} & ( & 3.2 & + & 10.6 & )\\
    \cline{2-36}
    & \textbf{MinkNet\_0.5$\times$~\cite{choy2019minkowski}} & 71.2 & \textbf{47} & ( & 8 & + & 39 & ) & \textbf{142} & ( & 23 & + & 119 & ) & \textbf{126} & ( & 33 & + & 93 & ) & 101 & 128 & 128 & \textbf{7.4} & ( & 2.8 & + & 4.6 & ) & \textbf{9.6} & ( & 3.2 & + & 6.4 & )\\
    
    \cline{2-36}
    & \textbf{SN\_32$\times$2048~\cite{cortinhal2020salsanext}} & 68.2 & \textbf{13} & ( & 8 & + & 5 & ) & \textbf{50} & ( & 23 & + & 27 & ) & \textbf{111} & ( & 41 & + & 70 & ) & 329 & 317 & 196 & \textbf{21.7} & ( & 3.4 & + & 18.3 & ) & \textbf{21.1} & ( & 3.2 & + & 17.9 & )\\
    \cline{2-36}
    & \textbf{SN\_32$\times$1024~\cite{cortinhal2020salsanext}} & 69.2 & \textbf{13} & ( & 8 & + & 5 & ) & \textbf{48} & ( & 26 & + & 22 & ) & \textbf{74} & ( & 36 & + & 38 & ) & 180 & 240 & 113 & \textbf{17.5} & ( & 3.2 & + & 14.3 & ) & \textbf{17.6} & ( & 3.5 & + & 14.1 & )\\
    
    \cline{2-36}
    & \textbf{SPVCNN~\cite{tang2020spvnas}} & 72.6 & \textbf{57} & ( & 8 & + & 49 & ) & \textbf{169} & ( & 22 & + & 147 & ) & \textbf{172} & ( & 31 & + & 141 & ) & 356 & 383 & 383 & \textbf{12.4} & ( & 2.7 & + & 9.7 & ) & \textbf{13.3} & ( & 3.0 & + & 10.3 & )\\
    \cline{2-36}
    & \textbf{SPVCNN\_0.5$\times$~\cite{tang2020spvnas}} & 69.6 & \textbf{58} & ( & 8 & + & 50 & ) & \textbf{169} & ( & 23 & + & 146 & ) & \textbf{157} & ( & 33 & + & 124 & ) & 122 & 149 & 149 & \textbf{10.3} & ( & 2.8 & + & 7.5 & ) & \textbf{10.0} & ( & 3.0 & + & 7.0 & )\\

    \cline{2-36}
    & \textbf{WI-48-256~\cite{puy2023waffleiron}} & 76.1 & \textbf{111} & ( & 64 & + & 47 & ) & \textbf{527} & ( & 114 & + & 413 & ) & \textbf{820}   & ( & 157 & + & 663 & ) & 426 & 750 & 747 & \textbf{19.9} & ( & 2.1 & + & 17.8 & ) & \textbf{16.2} & ( & 1.8 & + & 14.4 & )\\
    \cline{2-36}
    & \textbf{WI-12-128~\cite{puy2023waffleiron}} & 74.1 & \textbf{73} & ( & 59 & + & 14 & ) & \textbf{181} & ( & 99 & + & 82 & ) & \textbf{264} & ( & 157 & + & 107 & ) & 186 & 349 & 349 & \textbf{12.9} & ( & 2.8 & + & 10.1 & ) & \textbf{11.4} & ( & 2.9 & + & 8.5 & )\\

    
    \hline

  \end{tabular}
  }
\end{table*}

\subsection{Model-specific Analysis}
\begin{figure}[t!]
  \centering
    \includegraphics[width=0.475\textwidth]{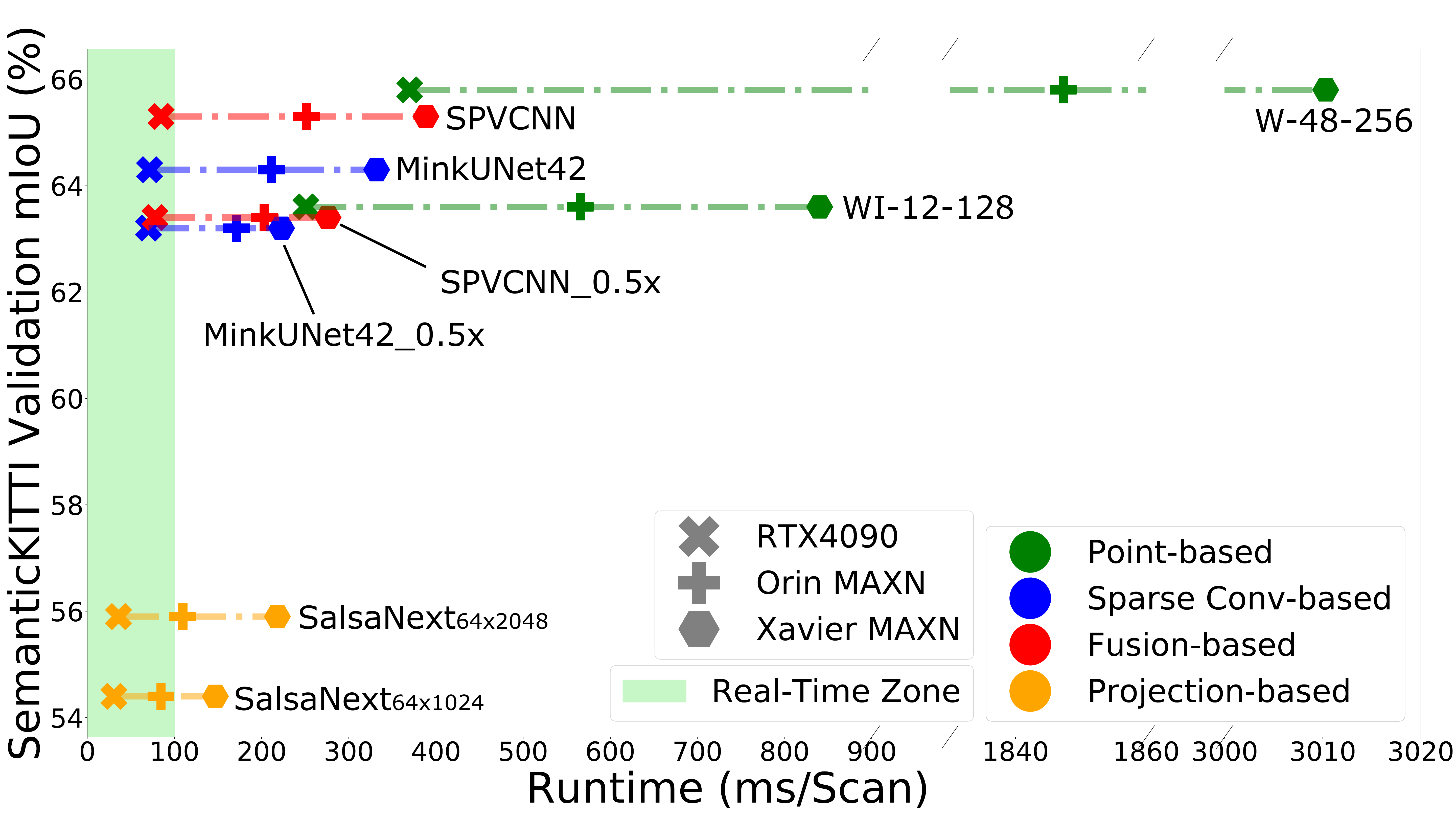}
    \caption{mIoU vs. runtime SemanticKITTI (10Hz input sensor)}
    \label{fig:semantickitti}
\end{figure}

\textbf{Minkowski and SPVCNN.} Both pre-process input data similarly, by applying a GPU sparse tensor quantization that generates sparse tensors and converts the input into unique hash keys and associated features. This proves to be the fastest Pre-Processing in \cref{tab:performance_metrics}. Further, MinkNet Inference time on a single validation scan of SemanticKITTI takes on average 166 ms and 266 ms on the AGX Orin and AGX Xavier, respectively. As for SPVCNN, the added point-wise MLP has an added overhead on the Inference of 43 ms and 63 ms on the Jetson platforms, respectively, which is a non-satisfactory trade-off with the overall accuracy gain of 1 mIoU on SemanticKITTI. Moreover, Minkowski achieves better accuracy than SPVCNN on nuScenes with faster Inference across all platforms. As illustrated in \cref{fig:nuscenes}, and \cref{fig:semantickitti}, both models remain far from real-time execution on embedded systems. Pruning 50\% channels for each sparse convolution layer leads to a 75\% reduction in \#Params and \#MACs (\cref{tab:complexity}),  but only slightly improves Inference time while significantly decreasing the mIoU. One approach to accelerate Inference is by reducing the total number of convolution layers, though this lowers the network's potential for accurate predictions. To counteract accuracy loss, substantial design modifications are needed for sparse convolutions and the network overall, including adjustments to kernel size and shape, voxels resolution, and channels dimension.




\textbf{SalsaNext.} This method achieves the fastest Inference time, as shown in \cref{fig:nuscenes} and \cref{fig:semantickitti}, with acceptable peak GPU memory consumption but high power consumption (see \cref{tab:performance_metrics}). Projecting the 3D point cloud to a LiDAR image requires significantly more CPU power than data Pre-Processing in other methods. Reducing the projection image width decreases total \#MACs (see \cref{tab:complexity}) and accelerates Inference time by about 50\% on SemanticKITTI, but results in a 1.5\% decrease in mIoU. Conversely, smaller LiDAR image sizes benefit nuScenes due to fewer points per scan. However, SalsaNext often shows lower mIoU accuracy due to the loss of 3D spatial features in projection. Additionally, back-projecting to the original point cloud resolution using K-nearest neighbors in Post-Processing leads to incorrect class predictions for several neighboring points. Despite being closest to real-time execution on embedded systems, SalsaNext's efficiency is accompanied by a reduced accuracy.


\textbf{WaffleIron.} 
\cref{tab:performance_metrics} shows that WaffleIron models demonstrate strong qualitative performance on SemanticKITTI and nuScenes but have the highest total runtime across all computing platforms and datasets. As illustrated in \cref{fig:nuscenes}, and \cref{fig:semantickitti}, WaffleIron is significantly far from real-time execution compared to other methods. Moreorver, it has the highest peak GPU memory usage, particularly for WI-48-256, but remains tolerated. Furthermore, WI-48-256 ranks among the highest in power consumption on the AGX Orin, especially for the GPU part. To reduce runtime and power consumption, we implement WI-12-128, which achieves over a 90\% reduction in \#Params and \#MACs (see \cref{tab:complexity}) with only a moderate drop in mIoU. Consequently, WI-12-128 offers faster Inference time, lower total power consumption, and less than half the GPU memory consumption. However, its Pre-Processing time remains significant, primarily dependent on the number of points per scan. The main components of WaffleIron Pre-Processing that accumulate in time are obtaining the index of corresponding 2D cell for each point on the projected grid and the nearest neighbors search, including constructing a kd-tree and querying neighbors essential for the embedding layer, and for subsequent upsampling. To accelerate Pre-Processing, parallelizing these tasks on available CPU workers and finding alternatives to neighborhood searches for point token generation are potential solutions. Further improvements to the backbone design are necessary to reduce the Inference time; otherwise, WaffleIron cannot be embedded with real-time consideration.

We extend our investigation in point-based methods to study a recent transformer-based architecture PTv3~\cite{wu2024ptv3}. It uses serialized neighbor mapping instead of KNN for receptive field expansion and improves processing speed by 3$\times$ from PTv2~\cite{wu2022ptv2}. In \cref{tab:ptv3}, we compare it to WI-48-256 on nuScenes, and it shows higher accuracy, comparable Inference, but 3× higher Pre-Processing time. Pruning 50\% channels of its attention layers, PTv3\_0.5$\times$ results in a slightly lower runtime but maintains a 2.5× higher Pre-Processing than WI-48-256 due to point cloud serialization. This implies that PTv3 struggles for real-time execution, and therefore is not investigated on Jetson embedded systems.

\begin{table}[t!]
\captionsetup{skip=0.1\baselineskip} 
\vspace{0.2cm}
  \caption{PTv3 performance on NVIDIA RTX4090 GPU.}
  \label{tab:ptv3}
    \centering
    \resizebox{0.47\textwidth}{!}{
    \begin{tabular}{@{}c@{}|@{}c@{\hskip 10pt}r@{\hskip 6pt}r@{\hskip 6pt}r@{\hskip 6pt}r@{}}
        \toprule
        \multicolumn{2}{l}{\makecell[l]{\textbf{Point-based (nuScenes)}}} & \multicolumn{2}{c}{\textbf{Complexity}} & \multicolumn{2}{c}{\textbf{Runtime}} \\
         \cmidrule(lr){3-4} \cmidrule(l){5-6}
         \textbf{Methods} & \textbf{mIoU}& \textbf{Params} & \textbf{MACs} & \textbf{Pre-Proc} & \textbf{Inference} \\
        \midrule
        WI-48-256~\cite{puy2023waffleiron}  & 76.1 & \hspace{5pt}6.8M & 122.4G &  64 ms & 47 ms \\
        \midrule
        PTv3~\cite{wu2024ptv3} & 78.5 & 15.3M & 96.9G & 198 ms & 43 ms \\
        PTv3\_0.5$\times$~\cite{wu2024ptv3}  & 77.5 & 3.9M & 24.3G & 166 ms & 35 ms \\
        \bottomrule
    \end{tabular}
    }
\end{table}

\section{CONCLUSIONS}
\label{sec:conclusion}
In this work, we benchmark various 3D neural networks for outdoor semantic segmentation on Jetson embedded platforms. Our assessment involves analysis on various performance metrics, and the findings indicate the need of substantial adjustments in network designs to target embedded systems. Our contribution lies in the critical evaluation of various methodologies, drawing attention to the inherent impracticalities. Given that a majority of robots and autonomous vehicles rely on embedded platforms as their core systems, it is noteworthy that prevailing methodologies in 3D perception are not predominantly tailored for deployment on such platforms. Our analysis exposes the limitations of these methodologies, pinpointing bottlenecks and advocating for substantial redesigns. Notably, we underscore the significance of Pre-Processing time, an aspect often overlooked by existing methods, highlighting its potential to be more critical than the Inference time itself. Particularly, our findings on point-based methods reveal a lack in their real-time capability even on powerful hardware, mainly due to heavy Pre-Processing. Sparse convolution-based demonstrate real-time performance on high-end GPUs but not on embedded systems. Projection-based methods prove to be the most efficient on embedded systems but at the cost of lower segmentation accuracy.
Thus, real-time and embedded LiDAR semantic segmentation remains a challenge for autonomous systems.


\addtolength{\textheight}{-12cm}   

\bibliographystyle{IEEEtran} 
\bibliography{iros_main}

\end{document}